\documentclass{article}

\usepackage{PRIMEarxiv}

\usepackage[utf8]{inputenc} 
\usepackage[T1]{fontenc}    
\usepackage{hyperref}       
\usepackage{url}            
\usepackage{booktabs}       
\usepackage{amsfonts}       
\usepackage{nicefrac}       
\usepackage{microtype}      
\usepackage{lipsum}
\usepackage{fancyhdr}       
\usepackage{graphicx}       
\usepackage{amsmath}
\graphicspath{{media/}}     

\usepackage{algorithm}%
\usepackage[noend]{algpseudocode}%
\usepackage{bm}
\usepackage{multirow}
\title{Uncovering Latent Phase Structures and Branching Logic in Locomotion Policies: A Case Study on HalfCheetah
}

\author{
  Daisuke Yasui*, Toshitaka Matsuki, Hiroshi Sato \\
  Mathematics and Computer Science \\
  National Defense Academy of Japan \\
  Yokosuka, Japan\\
  \texttt{*ed24003@nda.ac.jp} \\
}

\begin{document}
\maketitle

\begin{abstract}
In locomotion control tasks, Deep Reinforcement Learning (DRL) has demonstrated high performance; however, the decision-making process of the learned policy remains a black box, making it difficult for humans to understand.
On the other hand, in periodic motions such as walking, it is well known that implicit motion phases exist, such as the stance phase and the swing phase.
Focusing on this point, this study hypothesizes that a policy trained for locomotion control may also represent a phase structure that is interpretable by humans.
To examine this hypothesis in a controlled setting, we consider a locomotion task that is amenable to observing whether a policy autonomously acquires temporally structured phases through interaction with the environment.
To verify this hypothesis, in the MuJoCo locomotion benchmark HalfCheetah-v5, the state transition sequences acquired by a policy trained for walking control through interaction with the environment were aggregated into semantic phases based on state similarity and consistency of subsequent transitions.
As a result, we demonstrated that the state sequences generated by the trained policy exhibit periodic phase transition structures as well as phase branching.
Furthermore, by approximating the states and actions corresponding to each semantic phase using Explainable Boosting Machines (EBMs), we analyzed phase-dependent decision making—namely, which state features the policy function attends to and how it controls action outputs in each phase.
These results suggest that neural network–based policies, which are often regarded as black boxes, can autonomously acquire interpretable phase structures and logical branching mechanisms. 
\end{abstract}


\section{Introduction}
    Deep Reinforcement Learning (DRL) has achieved remarkable performance improvements in continuous control tasks and has been shown to be effective for complex problems such as robotic locomotion and motion control~\cite{Lillicrap2015Continuous}.
    Through learning, DRL generates a policy function that receives a state and outputs an action, thereby performing control tasks.
    The learned policy function is generally represented as a deep neural network, and its decision-making process becomes a black box.
    In periodic motion tasks such as walking and running, implicit motion phases, such as stance phase and swing phase exist, and stable locomotion is realized through their repeated transitions in a fixed order, a fact widely recognized in biomechanics and robotics~\cite{Collins2005EfficientBiped,Hurmuzlu2004HybridDynamics,Winter2009Biomechanics}.

    Based on these observations, we hypothesize that a policy trained for locomotion tasks may also represent a phase structure that is interpretable by humans.
    To confirm and interpret the existence of such a phase structure, we identified latent semantic phases acquired by a policy function trained on the MuJoCo locomotion benchmark HalfCheetah-v5 and analyzed each phase in detail.
    
    In the experiments, we first embedded the state sequences obtained through the interaction between the policy and the environment into a low-dimensional space.
    We then performed clustering incorporating a temporal constraint that encourages consecutive states to be assigned to the same cluster, thereby identifying a temporal phase structure.
    A temporal phase structure refers to a structure in which similar postures and motion states repeatedly appear in a fixed order when the state sequence is observed the state sequence generated by a continuous control policy along the time axis.
    This enables the state sequences generated by the policy to be segmented and interpreted in meaningful phase units.
    Next, by employing an Explainable Boosting Machine (EBM)\cite{Nori2019InterpretML}, we generated an approximate model that represents actions in each phase as the sum of additive contributions of state features, thereby explicitly describing the action control rule for each phase.
    This makes it possible to analyze which state features the policy function mainly relies on to control action outputs in each phase.
    
    The main contribution of this study is the empirical demonstration that a policy trained for locomotion tasks internally acquires a quasi-periodic phase structure.
    In addition, we analyzed which state features primarily control action outputs in each phase and the factors that cause action branching, based on the contributions of state features to action outputs.
    This perspective suggests that policy functions, often regarded as black boxes, may in fact possess phase structures that are understandable to humans, and is expected to provide insights for the advancement of Explainable Reinforcement Learning (XRL).
    
    The remainder of this paper is organized as follows.
    
    \begin{itemize}
        \item The Related Work section outlines TD3, a representative reinforcement learning method for locomotion control, and studies on interpretability in reinforcement learning, and positions the present study.
        \item The Methodology section describes in detail the identification of phase structures and the method for explicitly modeling action generation rules for each phase.
        \item The Experiments and Discussion section demonstrates that a quasi-periodic phase structure is observed in the state sequences acquired through interaction with the environment by a policy trained in the MuJoCo locomotion environment HalfCheetah-v5, and provides a detailed analysis of the corresponding approximate models for each phase.
        \item The Conclusion summarizes this study and discusses its limitations and future directions.
    \end{itemize}
\section{Related Work}
\subsection{Reinforcement Learning Methods (Continuous Control and TD3)}
    Reinforcement Learning (RL) is a framework in which an agent learns a policy that maximizes cumulative rewards through interaction with an environment.
    In particular, for continuous control tasks involving continuous state and action spaces, Deep Reinforcement Learning (DRL) based on deep neural networks has been widely adopted~\cite{SuttonBarto2018,VanHasselt2016DoubleQ}.
    As representative methods, Deterministic Policy Gradient (DPG) and its deep extension, Deep Deterministic Policy Gradient (DDPG), have been proposed and have demonstrated strong performance in continuous action spaces~\cite{Lillicrap2015Continuous}.
    Subsequently, Twin Delayed Deep Deterministic Policy Gradient (TD3) was introduced to mitigate the training instability and overestimation bias of DDPG~\cite{Fujimoto2018TD3}.
    TD3 significantly improves training stability and performance through several techniques: (i) Clipped Double Q-learning using two Q-functions, (ii) delayed policy updates, and (iii) target policy smoothing.
    TD3 has shown excellent performance on continuous control benchmarks such as MuJoCo~\cite{Todorov2012MuJoCo} and has been widely adopted as one of the standard methods for robotic locomotion and running control tasks~\cite{Peng2018DeepMimic,Todorov2012MuJoCo}.
    In these methods, the policy function is typically represented as a multilayer neural network that directly outputs continuous actions from input states.
    While neural networks possess high representational capacity, their internal decision-making processes become black boxes, making it difficult for humans to understand which state features contribute to action generation and why.
    In particular, in tasks such as locomotion where temporally continuous motion is essential, explaining policy behavior solely through input–output relationships at a single time step is insufficient.
    
\subsection{Explainable Reinforcement Learning (XRL)}
    Against this background, research on Explainable Reinforcement Learning (XRL), which aims to make the internal structures and decision-making processes of reinforcement learning agents understandable to humans, has been actively pursued~\cite{ChengYuXing2025SurveyXRL,Hickling2022ExplainabilityDRL}.
    In recent years, post-hoc explanation methods that seek to interpret the internal mechanisms of trained agents without degrading their performance have attracted particular attention.
    A representative approach in XRL is to visualize the contribution of state features to actions or value functions based on gradient information or sensitivity analysis.
    This includes the application of XAI techniques such as input gradients~\cite{Simonyan2014Saliency}, Integrated Gradients~\cite{Sundararajan2017IG}, and Layer-wise Relevance Propagation (LRP)~\cite{Bach2015LRP} to reinforcement learning~\cite{Greydanus2018VisualizingRL,Zahavy2016Graying,Mott2019Attention}.
    Although these methods are effective in revealing ``which state features contributed to action selection'' at a single time step, they have difficulty capturing policies as temporally continuous action sequences.
    On the other hand, there also exist studies that attempt to understand higher-level behavioral modes and structures of agents by analyzing internal representations rather than limiting explanations to single-time-step contributions.
    Zahavy et al.~\cite{Zahavy2016Graying} embedded the internal state representations of reinforcement learning agents trained on Atari games into a low-dimensional space and applied clustering to visualize acquired behavioral modes and strategic state groups.
    This approach is important in demonstrating that semantically distinct behavioral states can exist within a black-box policy.
    However, the clustering approach did not explicitly consider temporal transition structures.
    Moreover, it was difficult to quantitatively explain which state features contributed to action generation within each cluster and for what reasons.
    
    In periodic tasks such as walking and running, implicit locomotion phases—such as stance and swing phases—exist, and the meaning of actions is strongly tied to preceding and subsequent state transitions~\cite{Collins2005EfficientBiped,Hurmuzlu2004HybridDynamics,Winter2009Biomechanics}.
    Therefore, explanations limited to a single time step are insufficient for fully understanding overall policy behavior.
    Alternatively, approaches that approximate policies or value functions using interpretable surrogate models have also been proposed.
    For example, some studies attempt to grasp the overall decision structure by approximating the relationship between states and actions using additive models~\cite{acero2024distillingreinforcementlearningpolicies}.
    However, because all time steps are handled collectively, such approaches cannot explicitly separate the multiple locomotion phases inherent in walking tasks, making it difficult to interpret how phases transition and what roles they play.
    Furthermore, there are studies that achieve phase-aware control by explicitly providing phase variables to the policy~\cite{Holden2017PFNN}, but these assume phase structures are embedded at the model design stage and do not aim to reveal post-hoc what behavioral structures a trained policy has autonomously acquired internally.

    In summary, existing post-hoc XRL methods face two main limitations:  
    (i) they tend to focus on single-time-step contribution explanations, and  
    (ii) they are unable to simultaneously handle temporal phase structures and phase-dependent action generation rules.  
    The present study differs from these approaches in that it focuses on the phase structures that a locomotion policy is presumed to have internally acquired, and employs a method that enables integrated analysis of phase structures, phase transitions, and decision rationales within each phase.
\section{Analysis Method}\label{sec:Proposed_method}
\begin{figure}
    \centering
    \includegraphics[width=1\columnwidth,page=1, trim={50mm 55mm 70mm 35mm}, clip]{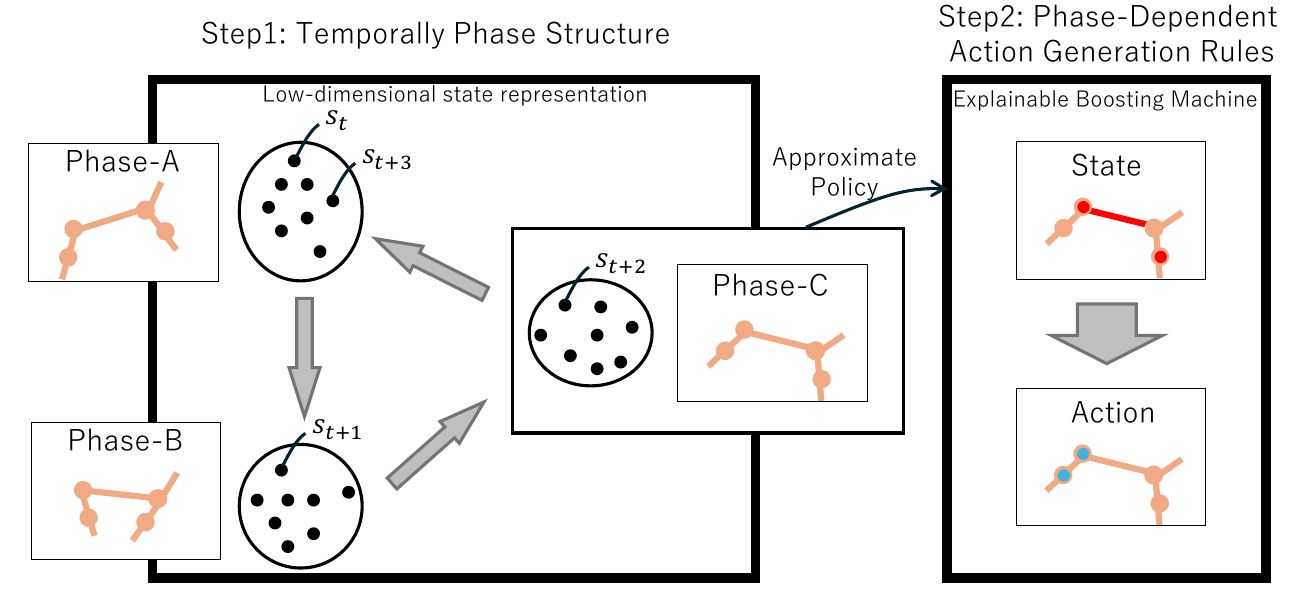}
    \vspace{-0.4cm}
    \caption{Overview of the analysis of a locomotion-control policy function. 
    This analysis is composed of two stages: 
    (Left) identification of temporally semantic phases by embedding state sequences and clustering with transition entropy minimization, and 
    (Right) phase-wise surrogate modeling using an Explainable Boosting Machine (EBM) to reveal state–action contribution rules within each phase.}
    \label{fig:proposed_method}
\end{figure}

The objective of the analysis conducted in this study is to clarify the phase structure that a locomotion control policy is presumed to have internally acquired, as well as the decision rationale underlying action generation within each phase.
The proposed analysis method is positioned as a post-hoc XRL approach that analyzes generated state–action sequences without retraining the learned policy.
As illustrated in Fig.~\ref{fig:proposed_method}, the analysis consists of the following two major stages:
\begin{itemize}
    \item Identification of the temporal phase structure
    \item Approximation of phase-wise action generation rules using an interpretable model
\end{itemize}

First, we focus on the state sequences generated through the interaction between the trained locomotion control policy and the environment, embed the state sequences into a low-dimensional space, and perform clustering.
By incorporating a temporal constraint such that the next state of states belonging to the same cluster is also likely to be assigned to the same cluster, we identify a temporal phase structure (Fig.~\ref{fig:proposed_method}, left).
A temporal phase structure refers to a structure in which clusters composed of states with similar state features—such as posture, joint angles, and velocities—appear repeatedly with stable ordering and transition rules rather than being randomly interchanged when the state sequence is observed along the time axis.

Next, for each identified phase, we approximate the actions generated by the policy function using an Explainable Boosting Machine (EBM) (Fig.~\ref{fig:proposed_method}, right).
This enables actions to be expressed as the sum of additive contributions of state features, thereby explicitly describing how much each state feature contributes to action control within each phase.
Each stage is described in detail below.

\subsection{State Embedding and Clustering for Phase Structure Identification}\label{sec:method_1}

The behavior of a policy in locomotion control tasks should be understood not only from a single time step but as a temporally continuous sequence of state transitions.
In particular, since phases such as stance and swing are expected to transition periodically in walking tasks, grouping similar states while considering temporal continuity may enable the identification of phase structures that are interpretable by humans.

First, we embed the state sequences obtained through the interaction between the trained policy function and the environment into a low-dimensional space using UMAP (Uniform Manifold Approximation and Projection)~\cite{McInnes2018UMAPJOSS}.
Through this process, semantically similar states are located closer together in the low-dimensional space.
Within this space, hierarchical clustering is performed for multiple numbers of clusters ($K=2$ to $20$), and the number of clusters $K$ that minimizes the conditional entropy $H_c$ across all clusters is selected so that temporal locomotion phases are aggregated into single clusters.

Dimensionality reduction via UMAP is applied prior to clustering because computing distances in the original state space may be susceptible to the curse of dimensionality; projecting onto a 2D manifold may allow the clustering algorithm to operate on a more geometrically meaningful representation.

UMAP embeds each state $s_t$ into a low-dimensional representation $z_t \in \mathbb{R}^{2}$ by minimizing the cross-entropy between the high-dimensional fuzzy graph weights $w_{ij}$ and their low-dimensional counterparts $v_{ij}$:
\begin{equation}
    \mathcal{L} = \sum_{(i,j)} \left[ w_{ij} \log \frac{w_{ij}}{v_{ij}} 
    + (1-w_{ij}) \log \frac{1-w_{ij}}{1-v_{ij}} \right],
\end{equation}
where the sum is taken over all pairs within $k_{\mathrm{umap}}$ nearest neighbors, $w_{ij} = \exp(-(d(s_i,s_j)-\rho_i)/\sigma_i)$ and $v_{ij} = (1+\|z_i-z_j\|^2)^{-1}$~\cite{McInnes2018UMAPJOSS}.
In this study, we set $k_{\mathrm{umap}}=15$ and minimum distance $= 0.1$.

The conditional entropy $H_c$ quantifies the uncertainty of transitions from a state belonging to one cluster to the cluster of the next state.
A smaller $H_c$ indicates that the next cluster is more predictable for states within a given cluster, implying that temporal locomotion phases are captured with high consistency.

First, each state at time step $t$ is assigned a cluster label $c_t \in \{1, \dots, K\}$ through clustering.
Let $N_i$ denote the number of states belonging to cluster $i$.
By aggregating cluster transitions along time series within the same episode, the number of transitions from cluster $i$ to cluster $j$, denoted as $N_{ij}$, is obtained.

Based on this, the transition probability distribution for cluster $i$ is estimated as
\begin{equation}
    p(c_{t+1}=j \mid c_t=i) = \frac{N_{ij}}{N_i}.
\end{equation}

The conditional entropy for cluster $i$ is given by

\begin{equation}
    H_i = - \sum_{j=1}^{K} p(c_{t+1}=j \mid c_t=i)
    \log p(c_{t+1}=j \mid c_t=i).
\end{equation}

Furthermore, to evaluate the overall consistency of phase transitions, we sum the conditional entropy 
$H_i$ of each cluster weighted by the occurrence frequency of states within that cluster
\begin{equation}
    H_c(K)
    = \sum_{i=1}^{K} p(c_t=i)\, H_i.
\end{equation}
where $p(c_t=i) = N_i / \sum_{k=1}^{K} N_k$ represents the probability that a state belongs to cluster $i$.
This $H_c(K)$ expresses how predictable the next phase is given the current phase, and smaller values indicate more deterministic and temporal phase transitions.
Within this low-dimensional space, we apply agglomerative hierarchical clustering using Ward's linkage criterion, which minimizes the increase in within-cluster variance at each merge step.
This tends to produce compact and homogeneous clusters, making it well-suited for identifying phases that reflect semantically coherent groups of states.

The optimal number of clusters $K^*$ is determined by evaluating $H_c(K)$ 
for each candidate $K \in \{2, 3, \ldots, 20\}$ and selecting the value 
that yields the global minimum:
\begin{equation}
    K^* = \arg\min_{\bm{K}} H_c(K).
\end{equation}

The obtained clusters are regarded as phases, and a phase transition graph is constructed by aggregating the transition relationships between phases.
This graph is a directed graph in which nodes represent phases and edges represent transition frequencies, enabling intuitive understanding of the periodicity and branching structures of motion internally acquired by the policy.

\subsection{Interpretation of Phase-wise Action Generation Rules (Approximation by EBM)}\label{sec:method_2}
After identifying the phase structure, we approximate the policy’s action generation mechanism within each phase using an Explainable Boosting Machine (EBM)~\cite{Nori2019InterpretML} in order to analyze it in detail.
EBM is a type of additive model that represents an action as follows:
\begin{equation}
    a = \sum_{i} f_i(s_i) + b,
\end{equation}
where $a$ denotes one component of the action vector, $s_i$ is a state feature, $f_i(\cdot)$ is the corresponding contribution function, and $b$ is a bias term.
Each contribution function $f_i(\cdot)$ is a one-dimensional nonlinear function approximated by a gradient-boosted decision tree ensemble.
This formulation enables independent approximation of the influence of each state feature on each action output.

In this study, state sequences are divided by phase, and an EBM is trained using only the state–action pairs corresponding to each phase to approximate the behavior of the policy function.
By observing the absolute values of the contribution scores of each state feature obtained from the EBM, excluding the bias term $b$, we focus on the magnitude of influence rather than the direction of increase or decrease, enabling us to examine which relationships between state features and action outputs are primarily emphasized by the policy within each phase.

Through this phase-wise approximation, it becomes possible to describe in a human-interpretable manner ``which actions are controlled based on which state features in which phase.''
Consequently, a black-box policy function can be understood as a structured action generation model composed of multiple phases and their transitions, enabling interpretation that captures temporal and structural aspects that conventional XRL methods have struggled to reveal.

\section{Experiments and Discussion}\label{sec:experiment}

We analyzed the phase structure that a locomotion-control policy is presumed to have internally acquired, as well as the decision rationale underlying action generation within each phase.
This section first describes the training setup and the target task, and then reports the results and discussion of (i)  ``identification of a temporal phase structure'' and (ii) ``phase-wise analysis using surrogate models'' in the corresponding subsections.

\begin{table}[htbp]
    \centering
    \caption{Experimental settings and specifications.}
    \label{tab:experimental_settings}
    \begin{tabular}{ll}
        \toprule
        \textbf{Item} & \textbf{Specification} \\
        \midrule
        Simulator & MuJoCo \\
        Benchmark Environment & HalfCheetah-v5 \\
        State Space Dimension & $17$ (Continuous) \\
        Action Space Dimension & $6$ (Continuous Torque) \\
        RL Algorithm & TD3 \\
        Policy Architecture & MLP (2 hidden layers with 256 units) \\
        Average Reward (over 5 eps.) & $9668.25 \pm 74.59$ \\
        \bottomrule
    \end{tabular}
\end{table}

In the experiments, we used the MuJoCo physics simulator~\cite{Todorov2012MuJoCo}, and adopted HalfCheetah-v5, a standard benchmark for walking and running tasks, as the evaluation environment (Table~\ref{tab:experimental_settings}).
In HalfCheetah-v5, the state space is a 17-dimensional continuous vector consisting of the robot posture and joint angles/angular velocities, and the action space is a 6-dimensional continuous torque vector corresponding to the joints.
The label names and value ranges of the state and action variables are described in the Appendix.
As the reinforcement learning algorithm, we employed Twin Delayed Deep Deterministic Policy Gradient (TD3)~\cite{Fujimoto2018TD3}, which is known for strong performance and training stability in continuous control.
The policy function is implemented as a fully connected neural network with two hidden layers of 256 units after the input layer, and outputs a 6-dimensional action at the final layer.
After sufficient training, the obtained policy achieved an average return of 9668.25 $\pm$ 90.3  over 5 episodes on HalfCheetah-v5, realizing stable running behaviors.
The original TD3 paper \cite{Fujimoto2018TD3} reports a score of approximately 10,000 on HalfCheetah; our result of 9,668 is thus consistent with the performance level of a well-trained TD3 policy, confirming that the analyzed policy has acquired stable and effective locomotion behavior.
In this study, we perform post-hoc analysis while keeping the trained policy function fixed.

\subsection{Identification of a Temporal Phase Structure}

\begin{figure}[H]
    \centering
    \includegraphics[width=0.9\columnwidth,page=2, trim={5mm 0mm 30mm 20mm}, clip]{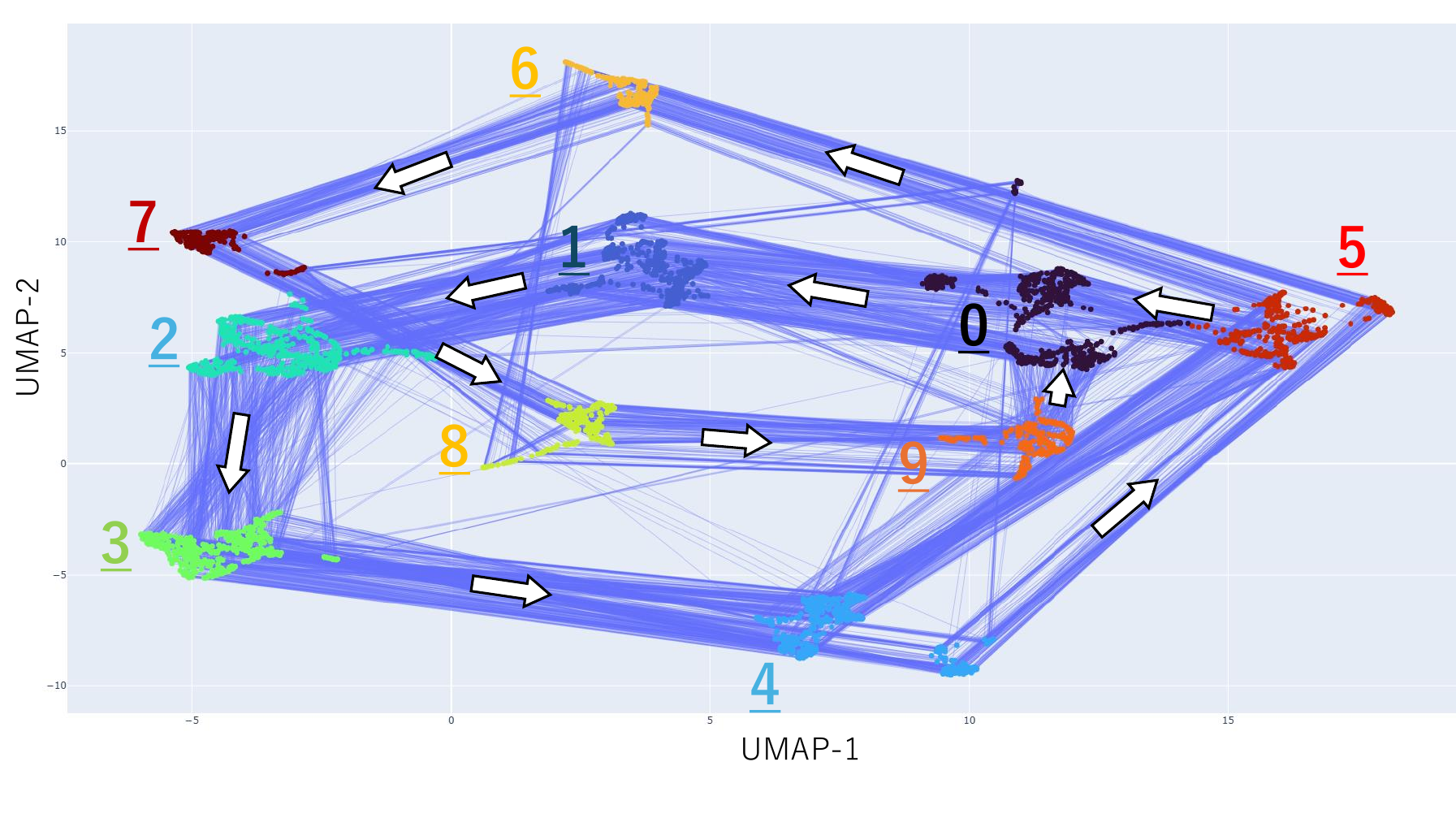}
    \vspace{-0.8cm}
    \caption{Visualization of the state space projected by UMAP. Each dot represents a state, colored by its assigned cluster (phase). 
    The blue edges connect each state to its successor state at the next time step. To clarify the dominant flow of phase transitions, white arrows are overlaid on the edges corresponding to transitions that cumulatively account for more than 70\% of observed transitions from each source cluster (see Table 2), indicating the direction of those transitions.}
    \label{fig:UMAP}
\end{figure}

We analyzed state sequences generated by the trained policy, consisting of 5 episodes $\times$ 1000 steps, and embedded them into a low-dimensional space based on the analysis method described in Sec.~\ref{sec:Proposed_method}.
After we projected the states into a 2D space using UMAP and performing hierarchical clustering so as to minimize the transition entropy, the number of clusters was determined as $K=10$.
Figure~\ref{fig:UMAP} visualizes the resulting clustering.
Each node represents a state in the 2D space; colors indicate clusters, and directed edges connect each state to its next state.
It can be visually observed that states in each cluster tend to transition to other clusters in a quasi-periodic manner.

In this subsection, we first quantitatively analyze the inter-cluster transition structure according to Sec.~\ref{sec:method_1} to verify whether each cluster functions as a temporally stable transition unit.
We then visualize the actual agent states belonging to each cluster and examine whether they have consistent kinematic/dynamic meanings, thereby discussing whether these clusters can be interpreted not merely as sets of states but as semantic phases internally acquired by the policy.

\subsubsection{Verification of Semantic Phase Properties via Inter-Cluster Transition Structure}
We constructed a cluster transition matrix by aggregating transition frequencies between clusters, as shown in Table~\ref{tab:cluster_matrix}.
Here, $N$ denotes the total number of states belonging to each cluster, and each column indicates which cluster the state transitioned to at the next time step.
As a result, for the HalfCheetah-v5 policy, the next cluster was uniquely determined with probability greater than 70\% for all clusters except Cluster 5.
In contrast, for Cluster 5, transitions to Cluster 6 and Cluster 0 occurred with almost equal probability, revealing a branching structure that is not fixed to a single next cluster.

From this observation, we identify two dominant transition patterns:
a cycle $0 \rightarrow 1 \rightarrow 2 \rightarrow 3 \rightarrow 4 \rightarrow 5 \rightarrow 0$ (Pattern 1), and a longer cycle
$0 \rightarrow 1 \rightarrow 2 \rightarrow 3 \rightarrow 4 \rightarrow 5 \rightarrow 6 \rightarrow 7 \rightarrow 8 \rightarrow 9 \rightarrow 0$ (Pattern 2).
These results suggest that the policy function does not merely repeat a simple periodic motion, but internally possesses a flexible control structure that branches into different action sequences depending on specific state conditions.

Overall, these results indicate that each cluster functions as a stable temporal unit with a limited number of likely successors, and the state sequences acquired through policy--environment interactions can potentially be interpreted as coherent groups of semantic phases.
\begin{table}[h]
    \centering
    \caption{Transition Matrix of Source vs. Target Clusters. Entries corresponding to destination clusters that cumulatively account for more than 70\% of transitions from each source cluster are shown in bold.}
    \label{tab:cluster_matrix}
    \setlength{\tabcolsep}{4pt} 
    \begin{tabular}{ccr rrrrrrrrrr}
        \toprule
        \multicolumn{2}{c}{\multirow{2}{*}{}} & \multirow{2}{*}{\textbf{N}} & \multicolumn{10}{c}{\textbf{Target Cluster}} \\ 
        \cmidrule(lr){4-13} 
        \multicolumn{2}{c}{} & & 0 & 1 & 2 & 3 & 4 & 5 & 6 & 7 & 8 & 9 \\ 
        \midrule
        
        \multirow{10}{*}{\rotatebox{90}{\textbf{Source Cluster}}} 
          & 0 & 764 & 137 & \textbf{563} & 3 & 0 & 0 & 0 & 16 & 44 & 0 & 0 \\
          & 1 & 622 & 0 & 5 & \textbf{610} & 5 & 0 & 0 & 0 & 1 & 0 & 0 \\
          & 2 & 645 & 6 & 0 & 20 & \textbf{612} & 5 & 0 & 0 & 0 & 2 & 0 \\
          & 3 & 621 & 1 & 0 & 0 & 3 & \textbf{609} & 7 & 0 & 0 & 0 & 0 \\
          & 4 & 615 & 77 & 0 & 0 & 0 & 0 & \textbf{537} & 0 & 0 & 0 & 0 \\
          & 5 & 546 & \textbf{276} & 22 & 1 & 0 & 0 & 2 & \textbf{244} & 0 & 0 & 0 \\
          & 6 & 260 & 0 & 0 & 6 & 0 & 0 & 0 & 0 & \textbf{206} & 48 & 0 \\
          & 7 & 251 & 0 & 0 & 0 & 1 & 0 & 0 & 0 & 0 & \textbf{250} & 0 \\
          & 8 & 324 & 0 & 0 & 0 & 0 & 1 & 0 & 0 & 0 & 24 & \textbf{299} \\
          & 9 & 352 & \textbf{267} & 32 & 0 & 0 & 0 & 0 & 0 & 0 & 0 & 53 \\ 
        \bottomrule
    \end{tabular}
\end{table}

\subsubsection{Verification of Semantic Phase Properties via Within-Cluster States}\label{sec:Step1}
\begin{figure}[H]
    \centering
    \includegraphics[width=1\columnwidth,page=23, trim={35mm 5mm 45mm 0mm}, clip]{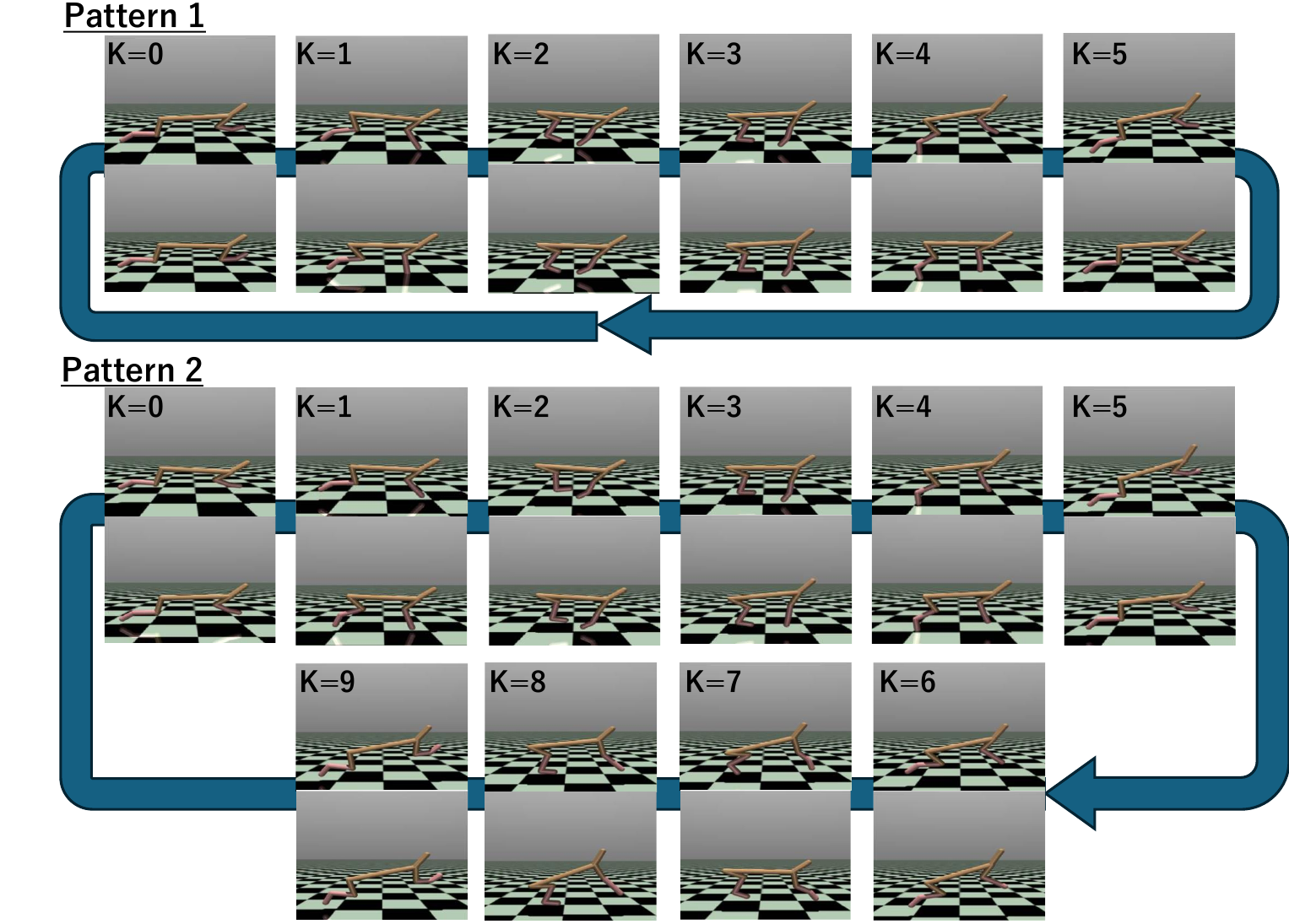}
    \vspace{-0.8cm}
    \caption{Randomly sampled rendered states from two phase--transition sequences.
    The upper row shows examples drawn from the transition cycle
    $\mathrm{Cluster}\;0 \rightarrow 1 \rightarrow 2 \rightarrow 3 \rightarrow 4 \rightarrow 5 \rightarrow 0$ (Pattern 1),
    while the lower row shows examples drawn from the longer cycle
    $\mathrm{Cluster}\;0 \rightarrow 1 \rightarrow 2 \rightarrow 3 \rightarrow 4 \rightarrow 5 \rightarrow 6 \rightarrow 7 \rightarrow 8 \rightarrow 9 \rightarrow 0$ (Pattern 2).
    For each cluster in the respective sequence, two states were randomly sampled
    from the corresponding state trajectory and rendered to visualize
    representative postures along the phase transitions.}
    \label{fig:3_random_states_in_each_cluster}
\end{figure}

Figure~\ref{fig:3_random_states_in_each_cluster} shows rendered states obtained by randomly sampling two trajectories each from state sequences corresponding to the two transition patterns identified in the previous subsection (Pattern 1 and Pattern 2).
In both patterns, the agent transitions from an airborne posture in Cluster 0 to front-leg ground contact in Cluster 1, pulls the rear leg in Clusters 2 and 3, leans backward in Cluster 4, and ascends into the air again in Cluster 5.
Compared with the case of transitioning from Cluster 5 to Cluster 0, the case of transitioning from Cluster 5 to Cluster 6 exhibits a more backward-leaning airborne posture.
The agent then lands with the rear leg (or both legs) in Cluster 7, ascends in Clusters 8 and 9, and returns to Cluster 0.
By repeatedly executing this sequence of locomotion phases, we can observe the overall mechanism of forward locomotion, and each cluster can be interpreted as representing a phase in the locomotion task.

\subsection{Phase-wise Analysis Using Surrogate Models}
Based on Sec.~\ref{sec:Step1}, we regard each identified ``cluster'' as a ``semantic phase,'' and hereafter simply refer to them as phases.

For each identified phase, we extracted state--action pairs and approximated the actions generated by the policy function using an Explainable Boosting Machine (EBM).
As a result, the coefficient of determination $R^2$ of the surrogate models for each phase ranged from approximately 0.6 to 0.9 (Table~\ref{tab:r2_score_split_booktabs}).
The lower $R^2$ in some phases (e.g., Phase 1) is attributed to the inability of EBMs, which are smoothed additive models, to fully capture high-frequency dynamics such as impact forces at ground contact.
However, from the perspective of our primary objective of interpretability, it is not strictly necessary to perfectly reproduce the torque values.
If the model captures the overall behavioral tendencies of the policy and the relative importance of which features it depends on, the interpretation can be considered qualitatively valid.

\begin{table}[htbp]
    \caption{Determination coefficient ($R^2$) of the EBM approximation for each phase.}
    \centering
    \label{tab:r2_score_split_booktabs}
    \resizebox{0.7\linewidth}{!}{
    \begin{tabular}{ccccccccccc}
        \toprule
        \textbf{K}   & 0 & 1 & 2 & 3 & 4 & 5 & 6 & 7 & 8 & 9 \\
        \midrule
        \textbf{$R^2$} & 0.76 & 0.61 & 0.70 & 0.65 & 0.72 & 0.90 & 0.68 & 0.68 & 0.85 & 0.82 \\
        \bottomrule
    \end{tabular}
    }
\end{table}

In this subsection, we first analyze which state features each phase-wise surrogate model primarily attends to and which action outputs it controls.
We then focus on Phase 5, whose successor phases split into two major branches, and separate the states that transition to Phase 0 from those that transition to Phase 6.
By comparing these cases, we discuss the policy's decision criteria for conditional branching.

\subsubsection{Interpretation of Phase-wise Action Generation Rules}
\begin{figure}[H]
    \centering
    \includegraphics[width=1\columnwidth,page=29, trim={20mm 45mm 30mm 45mm}, clip]{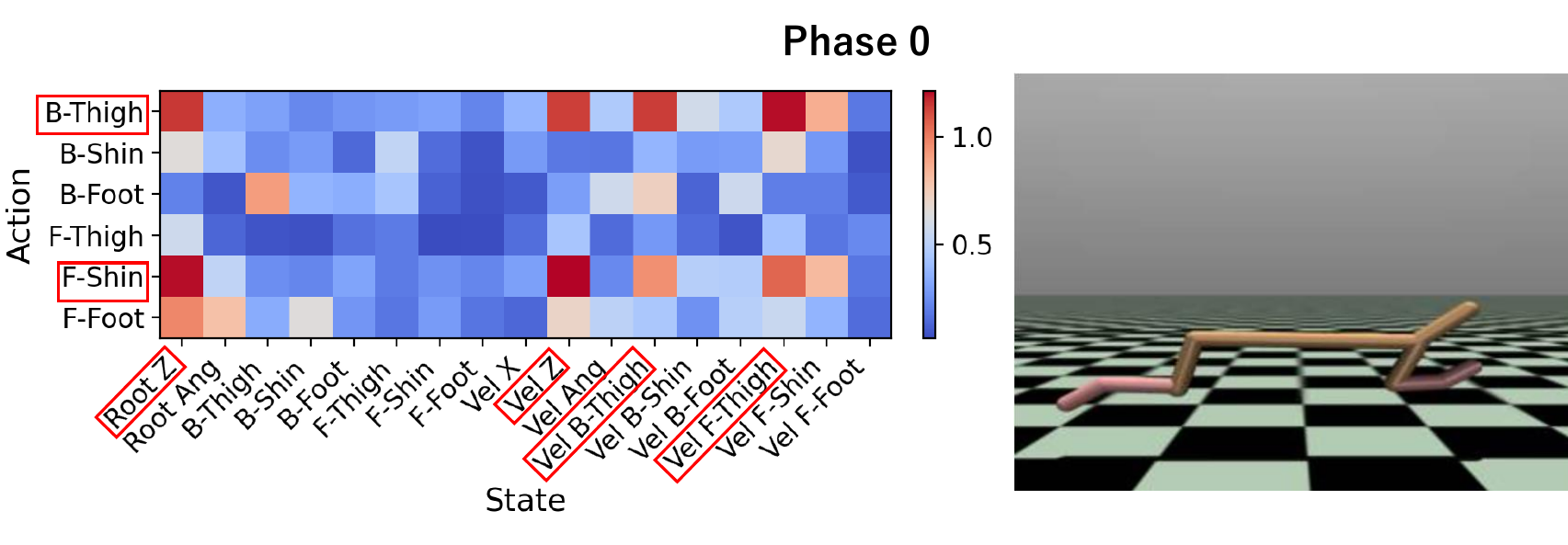}
    \vspace{-0.8cm}
    \caption{Feature attribution heatmaps for Phase 0 approximated by EBM, together with representative robot postures for each phase. Red frames indicate the state--action pairs whose attribution values exceed 95\% of the maximum heatmap value.}
    \label{fig:EBM_Heatmaps_0}
\end{figure}

\begin{figure}[H]
    \centering
    \includegraphics[width=1\columnwidth,page=37, trim={60mm 5mm 85mm 5mm}, clip]{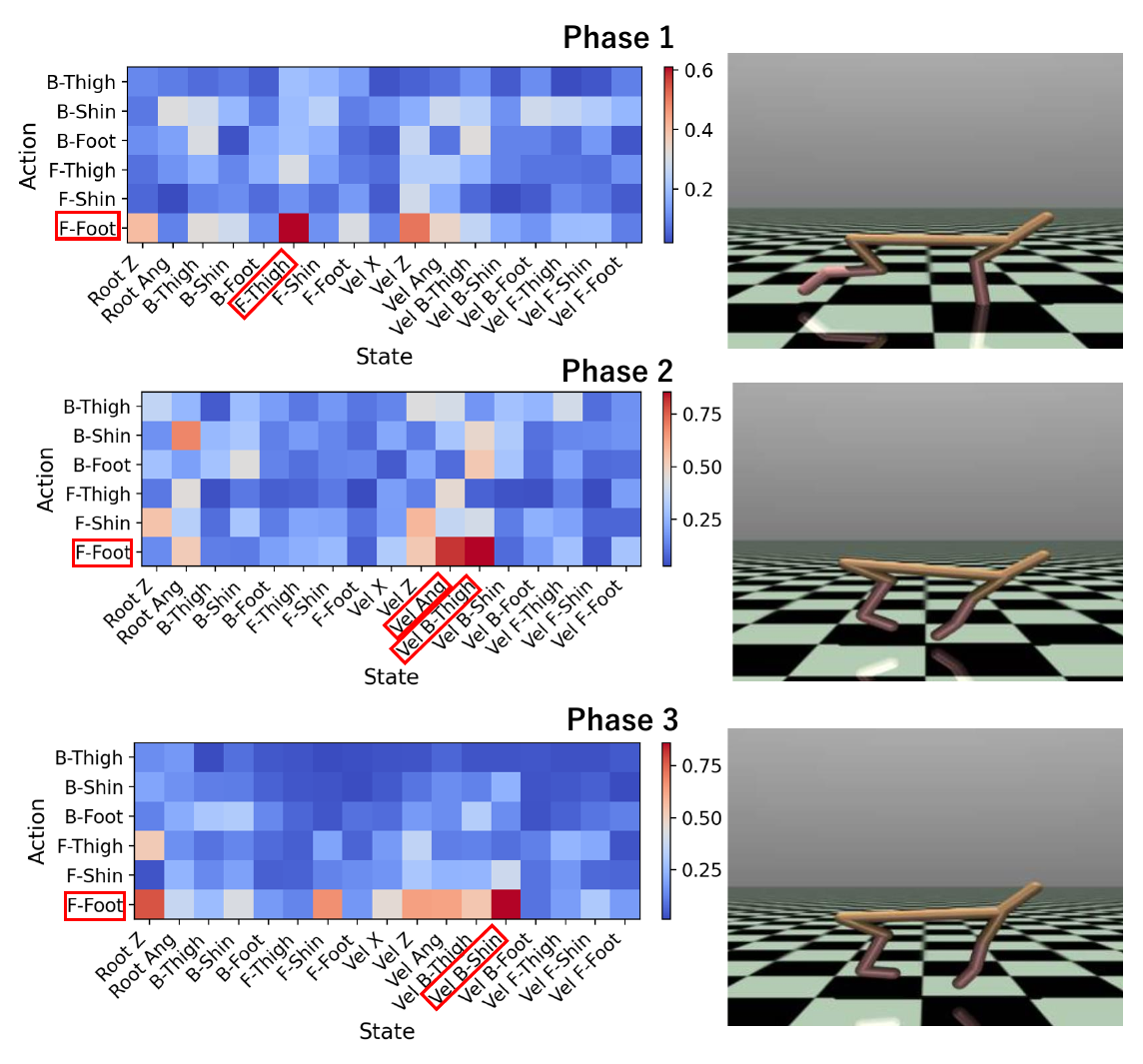}
    \vspace{-0.8cm}
    \caption{
    Feature attribution heatmaps for Phase 1, 2 and 3 approximated by EBM, together with representative robot postures for each phase. Red frames indicate the state--action pairs whose attribution values exceed 95\% of the maximum heatmap value.}
    \label{fig:EBM_Heatmaps_1_2_3}
\end{figure}

\begin{figure}[H]
    \centering
    \includegraphics[width=1\columnwidth,page=32, trim={20mm 0mm 30mm 0mm}, clip]{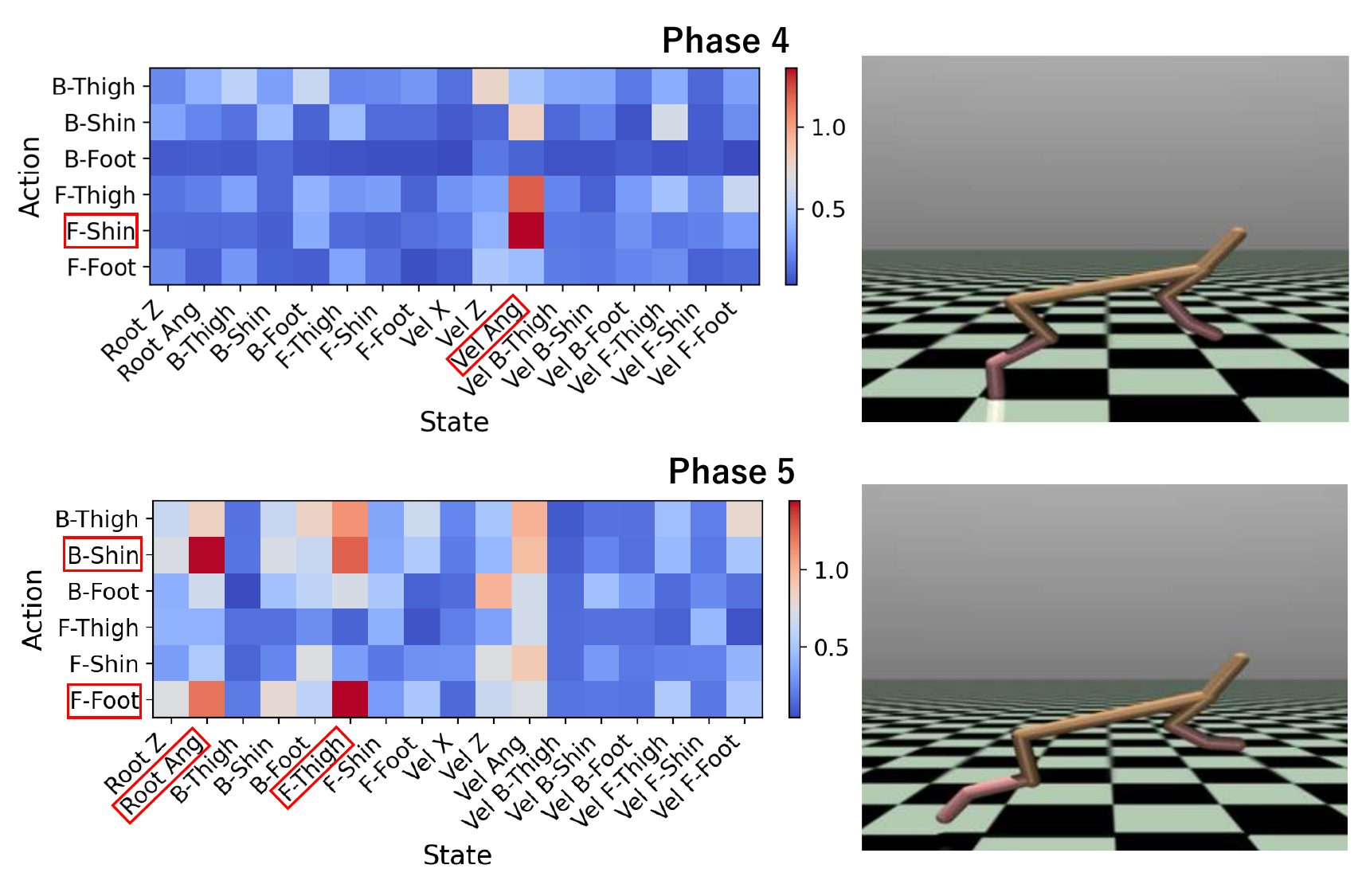}
    \vspace{-0.8cm}
    \caption{Feature attribution heatmaps for Phase 4 and 5 approximated by EBM, together with representative robot postures for each phase. Red frames indicate the state--action pairs whose attribution values exceed 95\% of the maximum heatmap value.}
    \label{fig:EBM_Heatmaps_4_5}
\end{figure}

\begin{figure}[H]
    \centering
    \includegraphics[width=1\columnwidth,page=33, trim={110mm 15mm 120mm 20mm}, clip]{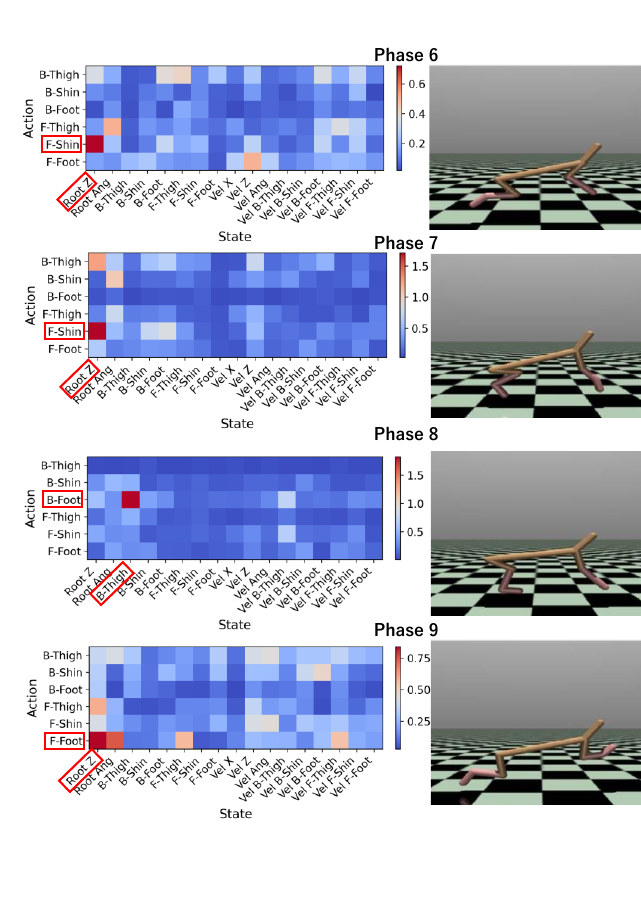}
    \vspace{-0.8cm}
    \caption{Feature attribution heatmaps for Phase 6 to 9 approximated by EBM, together with representative robot postures for each phase. Red frames indicate the state--action pairs whose attribution values exceed 95\% of the maximum heatmap value.}
    \label{fig:EBM_Heatmaps_6_9}
\end{figure}

By analyzing the EBMs trained to model the state--action relationship of the policy within each phase, we investigated which state features strongly contribute to which action outputs in each phase.
The left panels of Fig.~\ref{fig:EBM_Heatmaps_0} to Fig.~\ref{fig:EBM_Heatmaps_6_9} show heatmaps of the mean absolute contribution values to each action output over the state set of each phase, while the right panels show representative states for each phase.
Redder cells indicate state--action pairs with stronger influence, and the heatmaps are normalized by their minimum and maximum values.
Red boxes indicate state features and action outputs whose components fall within the top 5\% of the absolute contribution distribution.

Phase 0 corresponds to an airborne posture in which the agent’s limbs are widely extended (Fig.~\ref{fig:EBM_Heatmaps_0}).
The heatmap reveals that torso-height-related features such as \texttt{Root Z} and \texttt{Vel Z}, in addition to thigh angles, strongly influence joint control.
This suggests that the distance to the ground is the dominant information source for determining landing timing and posture control during airborne motion.

Phase 1 represents the moment when the front leg contacts the ground (Fig.~\ref{fig:EBM_Heatmaps_1_2_3}, top).
In addition to the height-related attention observed in Phase 0, the importance of \texttt{F-Thigh} increases and control of \texttt{F-Foot} becomes dominant.
This indicates that fine-grained torque control over the entire front leg is applied to achieve stable landing.
This result supports our claim that, even if the high-frequency torque spikes at impact are not perfectly reproduced, the EBM correctly captures the policy’s primary ``focus'' (i.e., the control of the front leg) in this phase.

Phases 2 and 3 depict the transition from front-leg support to rear-leg contraction 
(Fig.~\ref{fig:EBM_Heatmaps_1_2_3}, middle and bottom).
In Phase 2, pitch angular velocity (\texttt{Vel Ang}) and rear thigh velocity (\texttt{Vel B-Thigh}) strongly contribute to front-foot control.
In Phase 3, rear thigh velocity and torso height again play major roles.
These patterns suggest that, to prevent the body from tipping forward or backward after landing, the agent monitors body inclination, height, and rear-leg motion while controlling balance using the front leg.

In Phase 4, the agent leans backward, and in Phase 5 it transitions into an airborne state again (right side of Fig.~\ref{fig:EBM_Heatmaps_4_5}).
In Phase 4, \texttt{Vel Ang} primarily contributes to the control of \texttt{F-Shin}.
Because the body is tilted backward in this phase, it is inferred that the agent adjusts the front leg according to the body inclination to stabilize its posture.
In Phase 5, both the front and rear legs, such as \texttt{F-Foot} and \texttt{B-Shin}, are actively controlled.
As shown in Table~\ref{tab:cluster_matrix}, Phase 5 serves as a branching point toward Phases 0 and 6.
However, because the Phase 5 heatmap includes transitions to both Phase 0 and Phase 6, we further analyze Phase 5 in Sec.~\ref{sec:phase5_analyze} by separating the states that transition to Phase 0 from those that transition to Phase 6.
From Table~\ref{tab:cluster_matrix}, Phases 6 to 9 appear to form a cycle that occurs when the agent fails to transition from Phase 5 back to Phase 0.
In Phases 6 and 7, \texttt{Root Z} strongly influences the control of \texttt{F-Shin}, indicating posture adjustment based on height.
In Phase 8, information related to the rear thigh drives rear-foot control, leading to re-jumping toward Phase 9.
In Phase 9, torso height (\texttt{Root Z}) strongly affects front-foot control, suggesting height-based balance adjustment during airborne motion.

\subsubsection{Factor Analysis of Action Diversity in the Branching Phase}\label{sec:phase5_analyze}

\begin{figure}[H]
    \centering
    \includegraphics[width=0.8\columnwidth,page=35, trim={95mm 20mm 100mm 30mm}, clip]{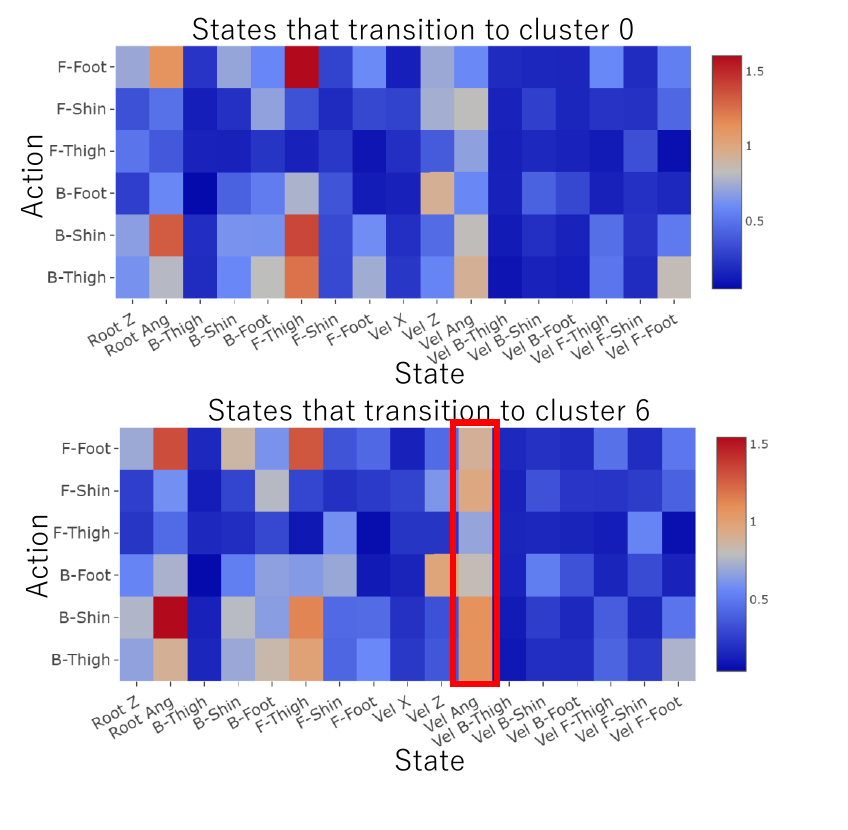}
    \vspace{-0.9cm}
    \caption{Comparison of feature attributions in the branching phase (Phase 5). The upper heatmap shows the contributions for transitions to Phase 0, while the lower heatmap shows the contributions for transitions to Phase 6. Note the distinct high contribution of Pitch Angular Velocity (Vel Ang) in the transition to Phase 6 (red box), indicating a momentum-based decision rule.}
    \label{fig:EBM_5}
\end{figure}

To analyze the branching mechanism in Phase 5, we compare heatmaps generated by training EBMs separately for the cases where Phase 5 transitions to Phase 0 and where it transitions to Phase 6.
As shown in Fig.~\ref{fig:EBM_5}, both heatmaps exhibit similar patterns in that they refer to \texttt{Root Ang} and \texttt{F-Thigh} while controlling \texttt{B-Shin} and \texttt{F-Foot}.
However, for transitions to Phase 6, the contribution of pitch angular velocity (\texttt{Vel Ang}) to joint torques becomes notably stronger (highlighted by red boxes).
This indicates that, in Phase 5, the policy dynamically switches its transition destination by adjusting control based on the momentum of pitch-angle changes during airborne motion.

This result strongly demonstrates the usefulness of the proposed method’s context-dependent interpretability.
Such a finding would likely be missed by conventional global importance analyses computed over entire episodes~\cite{acero2024distillingreinforcementlearningpolicies}.
This is because pitch angular velocity becomes important only within the short time window of Phase 5, while its importance is low in other phases; thus, its contribution would be diluted when averaged over all time steps.
By decomposing the time series into semantic phases as in our method, it becomes possible to identify the critical control logic that determines \emph{when} and \emph{what} triggers the agent to switch its decision making.
This capability is crucial for diagnosing the causes of unexpected behaviors and for verifying the safety of policies.

\section{Conclusion}
In this study, through experiments in the HalfCheetah-v5 environment, we provided empirical evidence that the trained policy function exhibits a periodic phase structure with branching, within which relatively simple action generation rules hold within each phase.
We also observed that, in phases where branching occurs, the policy may switch its behavior based on specific state features.

Although we conducted a detailed analysis on HalfCheetah-v5, where the periodic structure is evident, the applicability of the proposed analysis to non-periodic or event-driven tasks remains an important direction for future work.
In future research, we plan to evaluate the generality of this analysis method by applying it to other locomotion environments and different robot morphologies.

\section*{Acknowledgments}
This work was supported by JSPS KAKENHI Grant Numbers JP22K17969.

\section*{Data Availability and Access}
    Data that support the findings of this study are available from the corresponding author, Daisuke Yasui, upon reasonable request.
\section*{Appendix A: Details of the Experimental Environment}
\label{app:half_cheetah_details}

\begin{figure}[H]
    \centering
    \includegraphics[width=0.45\columnwidth,page=36, trim={95mm 70mm 115mm 30mm}, clip]{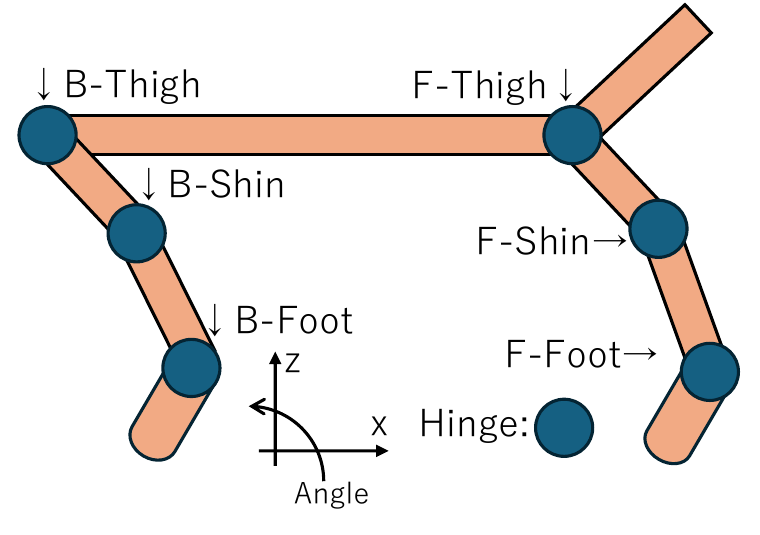}
    \caption{Actuated joints of the HalfCheetah robot. 
    The six joints used for control are the front and back thighs, shins, and feet.}
    \label{fig:halfcheetah_joints}
\end{figure}

In this study, we used the HalfCheetah-v5 environment provided by the Gymnasium library\cite{Todorov2012MuJoCo} as a benchmark task for reinforcement learning.
HalfCheetah is a two-dimensional cheetah-like robot model constructed on the MuJoCo physics simulator.
The robot consists of nine rigid body segments and eight joints, of which six joints (front and rear thighs, shins, and feet) are actuated by applying torques.
The objective of the task is to make the robot run forward (positive direction of the $x$-axis) as fast as possible.
The actuated joints used for control are illustrated in Fig.~\ref{fig:halfcheetah_joints}.

\subsection*{Observation Space}
The observation space $\mathcal{S}$ is a 17-dimensional continuous vector in $\mathbb{R}^{17}$ composed of generalized coordinates (positions and angles) and generalized velocities of the robot.
To preserve the Markov property while preventing overfitting to a specific absolute position, the robot’s absolute $x$ coordinate is excluded from the observation.
The detailed components of the observation vector are listed in Table~\ref{tab:half_cheetah_obs}.

\begin{table}[ht]
    \centering
    \caption{Details of the observation space in HalfCheetah-v5 (Dimension: 17)}
    \label{tab:half_cheetah_obs}
    \begin{tabular}{clll}
        \toprule
        Index & Variable Name & Physical Meaning & Unit \\
        \midrule
        0 & \texttt{Root Z} & Root (torso) $z$ position (height) & m \\
        1 & \texttt{Root Ang} & Root pitch angle & rad \\
        2 & \texttt{B-Thigh} & Back thigh angle & rad \\
        3 & \texttt{B-Shin} & Back shin angle & rad \\
        4 & \texttt{B-Foot} & Back foot angle & rad \\
        5 & \texttt{F-Thigh} & Front thigh angle & rad \\
        6 & \texttt{F-Shin} & Front shin angle & rad \\
        7 & \texttt{F-Foot} & Front foot angle & rad \\
        \midrule
        8 &  \texttt{Vel Root X} & Root velocity in $x$ direction & m/s \\
        9 &  \texttt{Vel Root Z} & Root velocity in $z$ direction & m/s \\
        10 & \texttt{Vel Root Ang} & Root angular velocity & rad/s \\
        11 & \texttt{Vel B-Thigh} & Back thigh angular velocity & rad/s \\
        12 & \texttt{Vel B-Shin} & Back shin angular velocity & rad/s \\
        13 & \texttt{Vel B-Foot} & Back foot angular velocity & rad/s \\
        14 & \texttt{Vel F-Thigh} & Front thigh angular velocity & rad/s \\
        15 & \texttt{Vel F-Shin} & Front shin angular velocity & rad/s \\
        16 & \texttt{Vel F-Foot} & Front foot angular velocity & rad/s \\
        \bottomrule
    \end{tabular}
\end{table}

\subsection*{Action Space}
The action space $\mathcal{A}$ is a 6-dimensional continuous vector in $\mathbb{R}^{6}$ representing the torques applied to each actuated joint.
Each element is normalized within the range $[-1.0, 1.0]$.
The correspondence between action vector elements and joints is summarized in Table~\ref{tab:half_cheetah_act}.

\begin{table}[ht]
    \centering
    \caption{Details of the action space in HalfCheetah-v5 (Dimension: 6, Range: $[-1, 1]$)}
    \label{tab:half_cheetah_act}
    \begin{tabular}{cll}
        \toprule
        Index & Joint Name & Controlled Target \\
        \midrule
        0 & \texttt{B-Thigh} & Torque applied to back thigh rotor \\
        1 & \texttt{B-Shin} & Torque applied to back shin rotor \\
        2 & \texttt{B-Foot} & Torque applied to back foot rotor \\
        3 & \texttt{F-Thigh} & Torque applied to front thigh rotor \\
        4 & \texttt{F-Shin} & Torque applied to front shin rotor \\
        5 & \texttt{F-Foot} & Torque applied to front foot rotor \\
        \bottomrule
    \end{tabular}
\end{table}

\bibliographystyle{plain}
\bibliography{sn-bibliography}

\end{document}